\documentclass{article}

\PassOptionsToPackage{numbers, compress}{natbib}

\usepackage[preprint]{neurips_data_2023}

\usepackage[utf8]{inputenc} 
\usepackage[T1]{fontenc}    
\usepackage{hyperref}       
\usepackage{url}            
\usepackage{booktabs}       
\usepackage{amsfonts}       
\usepackage{nicefrac}       
\usepackage{microtype}      
\usepackage{xcolor}         

\usepackage{graphicx}
\usepackage{amsmath,bm}
\usepackage{mathtools}
\usepackage{subfig}

\usepackage[noend]{algpseudocode}
\usepackage{algorithmicx,algorithm}

\usepackage{wrapfig}

\title{EVBattery: A Large-Scale Electric Vehicle Dataset for Battery Health and Capacity Estimation}

\author{
	Haowei He \thanks{Tsinghua University, hhw19@mails.tsinghua.edu.cn, jingzhaoz, wangyn7@mail.tsinghua.edu.cn, bbjiang, hanxuebing, guodongxu, ouymg@tsinghua.edu.cn}\\ Tsinghua IIIS
	\And
	Jingzhao Zhang\footnotemark[1] \\ Tsinghua IIIS
	\And
	Yanan Wang\footnotemark[1] \\ School of Vehicle and Mobility\\ Tsinghua University
	\And
	Benben Jiang\footnotemark[1] \\Department of Automation \\ Tsinghua University
	\And
	Shaobo Huang \thanks{ Circue Energy Technology Co., Ltd., huangshaobo, zhangyang, xionggengang@thinkenergy.net.cn} \\ Circue Energy Technology\\
	\And
	Chen Wang \thanks{School of Automation Science and Electrical Engineering, Beihang University, wangchen512@buaa.edu.cn}\\ Beihang University \\ 
	\And
	Yang Zhang \footnotemark[2]  \\  Circue Energy Technology
	\And
	Gengang Xiong \footnotemark[2]  \\  Circue Energy Technology
	\And
	Xuebing Han \footnotemark[1]\\ School of Vehicle and Mobility\\ Tsinghua University
	\And
	Dongxu Guo \footnotemark[1]\\ School of Vehicle and Mobility\\ Tsinghua University
	\And
	Guannan He \thanks{Department of Industrial Engineering and Management, Peking University, gnhe@pku.edu.cn}\\ Peking University
	\And
	Minggao Ouyang \footnotemark[1] \\ School of Vehicle and Mobility\\ Tsinghua University
}

\begin{document}
	
	\maketitle
	
	\begin{abstract}
		
		
		
		Electric vehicles (EVs) play an important role in reducing carbon emissions. As EV adoption accelerates, safety issues caused by EV batteries have become an important research topic. In order to benchmark and develop data-driven methods for this task, we introduce a large and comprehensive dataset of EV batteries. Our dataset includes charging records collected from hundreds of EVs from three manufacturers over several years. Our dataset is the first large-scale public dataset on real-world battery data, as existing data either include only several vehicles or is collected in the lab environment. 
		Meanwhile, our dataset features two types of labels, corresponding to two key tasks - battery health estimation and battery capacity estimation. In addition to demonstrating how existing deep learning algorithms can be applied to this task, we further develop an algorithm that exploits the data structure of battery systems. Our algorithm achieves better results and shows that a customized method can improve model performances. We hope that this public dataset provides valuable resources for researchers, policymakers, and industry professionals to better understand the dynamics of EV battery aging and support the transition toward a sustainable transportation system. 
	\end{abstract}
	
	\section{Introduction}
	
	

	In recent years, the popularity of electric vehicles (EVs) has significantly increased due to improved cruise range,  and reduced costs of onboard lithium-ion batteries~\citep{schmuch2018performance, mauler2021battery}. On the other hand, the high energy density and complex manufacturing process can also produce defective battery cells that have short life cycles or even lead to fire incidents. Although battery health and capacity can be estimated via chemical and physical tests, such a destructive method often renders the battery unusable and causes significant economic losses for users and automakers~\citep{rezvanizaniani2014review}. Monitoring the charging and discharging process of the EV battery system and analyzing battery performance and health through data analysis~\citep{goebel2008prognostics, widodo2011intelligent, li2020battery, xue2021fault, hong2019fault} has become a better choice as a gentle testing method.
	
	In addition to traditional mathematical statistical analysis, deep learning has achieved remarkable achievements in similar tasks. The success of neural network models has shown the huge demand for data quantity and quality~\citep{radford2019language}. However, the availability of public large-scale battery data for EVs is limited. Existing battery datasets often have limitations such as low sample size, lack of diversity,  or requiring synthetic simulations~\citep{li2020battery, hong2019fault, yang2020characterization}, which  limits the application of deep learning to real-world battery systems.
	
	To address this issue, we release a large-scale dataset of battery charging time-series data collected from EV charging stations. The dataset includes over 1.2 million charging snippets from 464 different EVs produced by three manufacturers. Each charging snippet contains 128 charging sampling time-series recording points. The data includes average cell voltage, charging current, temperature, battery capacity, estimated SOC during charging, and anomaly labels manually labeled afterward. This dataset aims to support deep learning research and development in EVs, including the evaluation of charging behavior, battery degradation, battery safety, and energy management systems, to promote research on improved performance, safety, and sustainability. 
	
	In this article, we explore two possible applications of this dataset - battery system health estimation and capacity estimation. We benchmark several machine learning and deep learning methods. Meanwhile, for the battery system health estimation task, we design the DyAD~(DYnamical system Anomaly Detection) algorithm based on the dynamic system characteristics of EV battery charging, achieving the best results in detecting battery health problems.
	
	The primary contributions of this paper are as follows:
	\begin{itemize}
		\item We provide a large-scale real-world dataset of EV batteries, including charging voltage, current, temperature, and estimated SOC. 
		\item The dataset has two types of labels -- battery health and battery capacity. We benchmark machine learning and deep learning algorithms in predicting both labels. The experimental results demonstrate that the dataset poses new  research challenges.
		\item For the battery health estimation task, we design the DyAD algorithm based on the dynamic system characteristics of EV battery charging, achieving the best results in the unsupervised anomaly detection task. This suggests that algorithms designed for the dataset can have better performance than general-purpose methods.
	\end{itemize}

	\section{Related works }

	To the best of our knowledge, there is no well-established deep-learning study of real-world battery systems with public large-scale datasets. 
	Existing related studies may not incorporate enough vehicles in their datasets, use lab experimental data, or use a closed-source dataset. We notice that \cite{li2020battery} collect data from nine vehicles and divide them into three categories. \cite{hong2019fault} record time-series information of one electric taxi. \cite{zheng2020micro} use four cells with inconsistent capacities. \cite{yang2020characterization} use eight battery cells to study faults in the battery pack. Other works use a larger but private dataset \citep{hong2019fault, li2020battery, yang2020characterization}.  
	
	The absence of large public datasets hinders the progress of deep learning techniques in this area. Many studies still use statistical methods~\citep{xue2021fault} or canonical machine learning algorithms~\citep{zheng2020micro} to study EV battery fault diagnosis and anomaly detection. LSTM-based deep neural networks are used in some time-series battery studies~\citep{hong2019fault, li2020battery}. However, the advanced time-series anomaly detection algorithms have already tuned into new techniques such as graph structure~\citep{GDN, MTADGAT} and achieved higher performance in the AI community. Whether these techniques generalize to battery time-series studies is still an open problem. Similarly, EV battery capacity studies have encountered the same problems~\citep{ng2020predicting, liu2020data}.

	\section{EVBattery dataset}
	\label{sec:evbattery_dataset}
	
	In this section, we first introduce the battery charging records, stored as time-series data, in our dataset. We will also briefly discuss the health and capacity labels, but will leave more details later in the next section.
	
	\paragraph{Data format} Our EVBattery dataset contains EV charging records collected from electric vehicle charging stations in collaboration with our research group. 
	The EVs are produced by three different manufacturers. 
	The raw charging data is divided into snippets using fixed-length sliding windows. Each charging snippet of length 128 consists of two parts: time-series sequences and meta-information. The time-series data comprises eight dimensions, including the average cell voltage, charging current, maximum and minimum cell voltage, maximum and minimum cell temperature, state of charge (SOC), and timestamp recorded during the charging process. The meta-information includes the vehicle number, current mileage, index of the snippet, battery health label, and current capacity label associated with the charging snippet. 
	
	The battery health label is determined based on fire incidents and lithium plating reports up until the collection date, and each charging snippet has its own health label. We classify vehicles that have such fault reports as anomalies, and label other vehicles as normal. 
	The battery capacity label is obtained by manufacturer engineers. The capacity label is a real number ranging from  28.28 to 46.23, where a higher value represents better health status. Due to the complexity of the charging process, only a subset of charging snippets can be marked with battery capacity by engineers.
	
	\paragraph{Data Anonymization}We named the three manufacturers as battery dataset 1, 2, and 3. The dataset only includes the average cell voltage without revealing the actual cell information. We perturbed and interpolated the original average voltage, current, and temperature values. Furthermore, we also randomly shifted and scaled the timestamps and mileages. Our test in Appendix~\ref{appendix_unanomymized_result} suggests that the anonymization does not hurt the test performance of our benchmarked methods.
	
	 \begin{figure}
		     \centering
		     \includegraphics[width=0.6\linewidth]{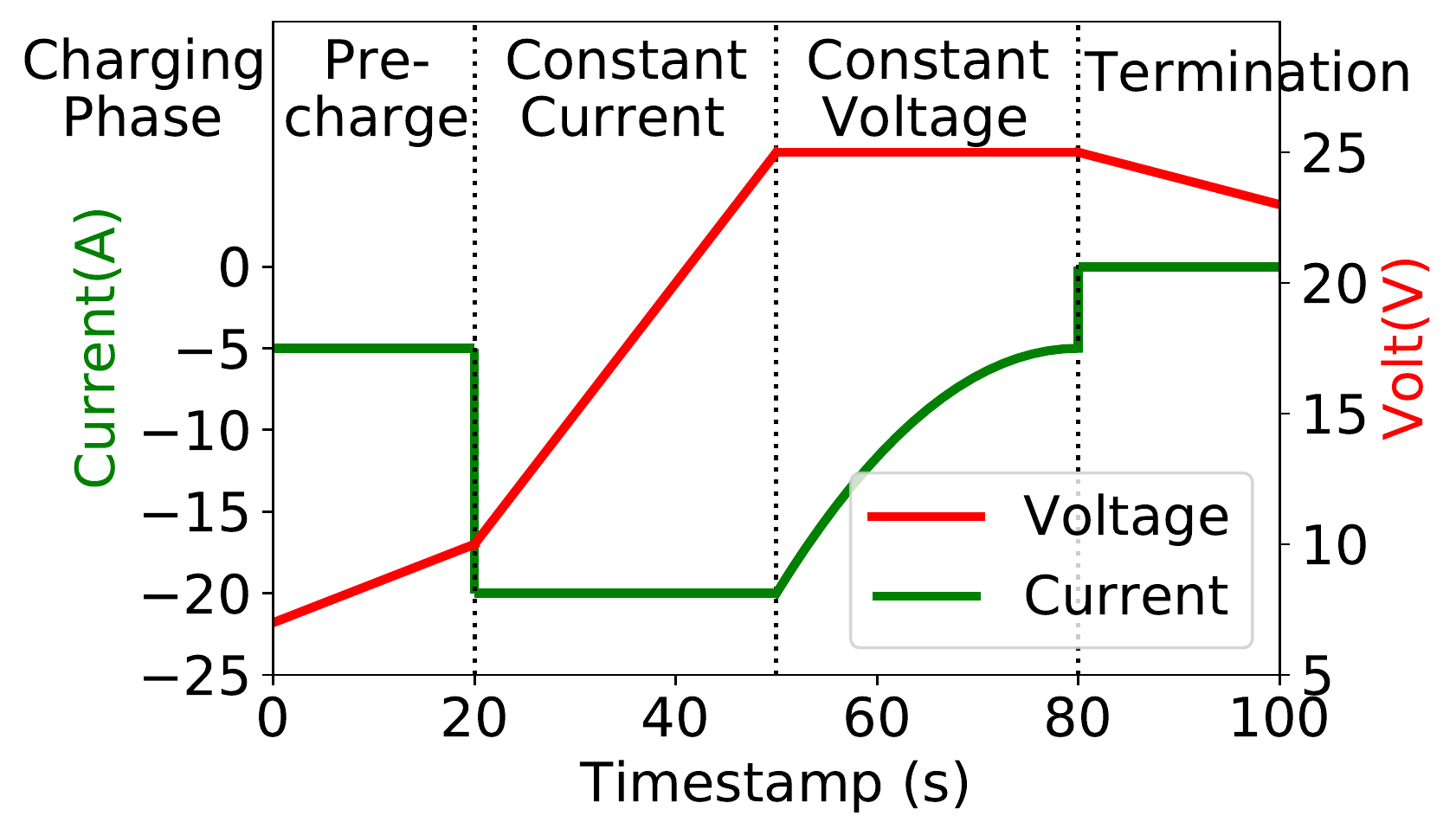}
		     \caption{Voltage and current changes in different stages of a charging process. }
		     \label{fig:charging_process1}
		 \end{figure}
	
	
	\paragraph{Battery charging patterns} To understand the complexity of the EV charging process, we first provide a brief introduction to the methods of EV charging. The complex charging pattern of recent advanced fast-charge batteries brings difficulties in analyzing the battery charging data. A typical complete charging process consists of three types of stages: pre-charging, constant-current charging, and constant-voltage charging. Pre-charging adjusts the battery state in preparation for high-current fast charging. Constant current charging can quickly store electrical energy. Constant voltage charging is the battery conditioning phase to maximize battery capacity. Here we draw a typical voltage and current curve during charging at different stages in Figure~\ref{fig:charging_process1}. More importantly, EV owners' behaviors, such as prematurely ending the charging process, can result in incomplete charging data, deviating from the pre-planned charging process. This in turn can pose challenges for data analysis.

\begin{figure}[h]
		\centering
		\includegraphics[width=0.8\linewidth]{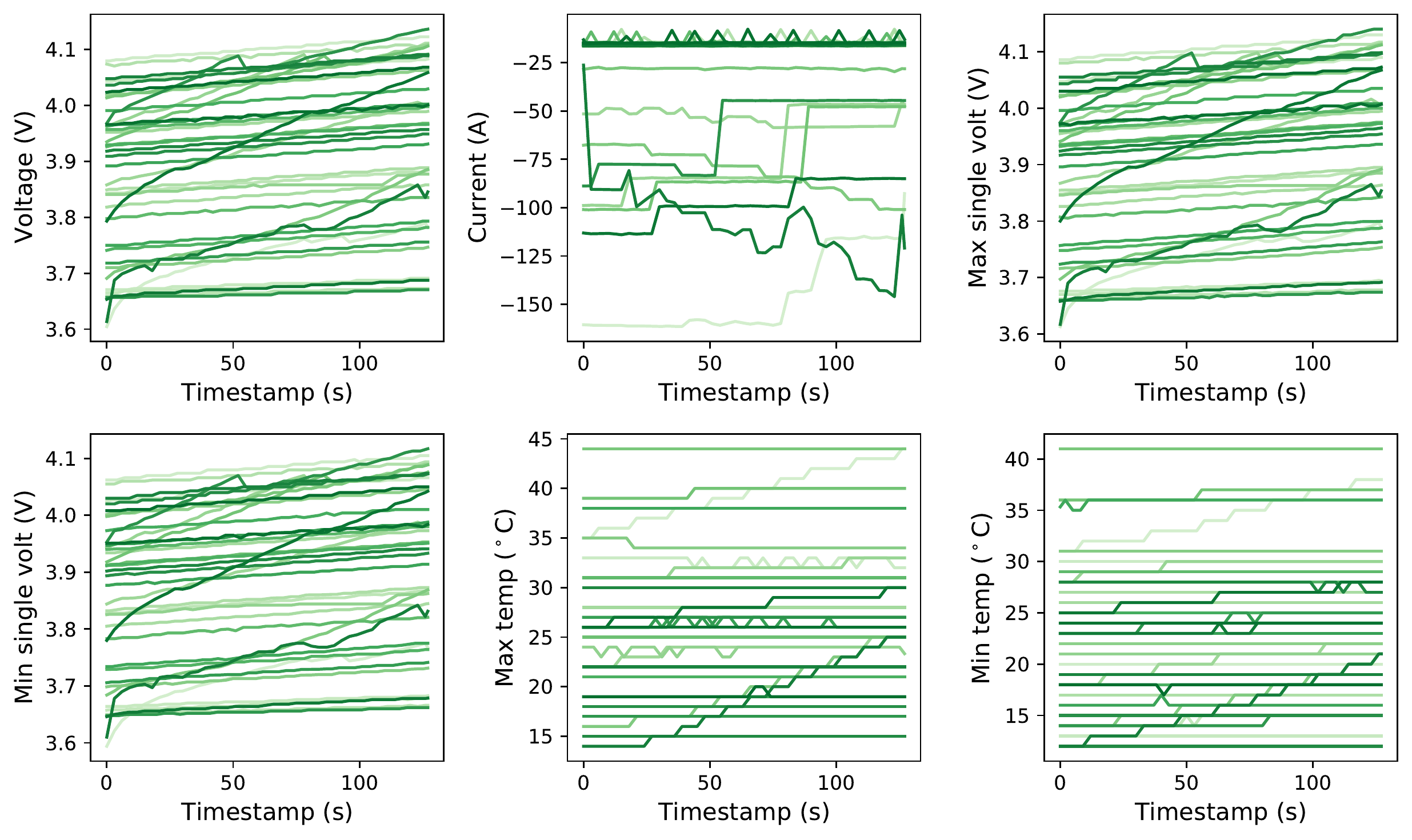}
		\caption{Sampled sequential data from a vehicle in our dataset. We can see these records essentially cover various charging scenarios, such as low temperature, high temperature, fast charging, slow charging, and transition of the charging stages. Darker lines represent earlier records. }
		\label{fig:dataset_example}
	\end{figure}

	\paragraph{Dataset examples and statistics} In Figure~\ref{fig:dataset_example}, we present some examples from the dataset to show its diversity. Detailed statistics about the data and label information are provided in Table~\ref{table:dataset_statistics}. Notice that the battery health labels~(anomaly labels) are assigned on a per-vehicle level and then extended to each charging snippet belonging to that vehicle. Additionally, due to technical limitations and challenges, engineers are only able to label the capacity tag for a subset of snippets.
	
	\begin{table}[h]
		\caption{\label{table:dataset_statistics}Statistics of our EVBattery dataset . Anomaly labels are obtained on the vehicle level and capacity labels are sparse among all snippets.}
		\begin{tabular}{cccc}
			\toprule
			& Battery dataset 1 & Battery dataset 2 & Battery dataset 3\\
			\midrule
			Number of vehicles                                                                & 217               & 198               & 49                \\
			Number of anomaly vehicles                                                        & 31                & 1                 & 16                \\
			Number of charging snippets                                                       & 629,121            & 472,829            & 176,327            \\
			\begin{tabular}[c]{@{}l@{}}Number of capacity labels\end{tabular} & 349,741            & 203,207            & 32,974  \\
			\bottomrule
		\end{tabular}
	\end{table}
	
	\paragraph{Dataset availability}
	Our dataset and code are available at \url{https://1drv.ms/f/s!AnE8BfHe3IOlg13v2ltV0eP1-AgP?e=9o4zgL}. The usage of our data and code is under a CC BY-NC-SA license. 
	We also hope our dataset will enable more applications for EVs. More information can be found in Appendix~\ref{appendix:more_dataset_information}.

	\section{Labels and tasks for the EVBattery dataset}

	Two important factors that determine a battery's value are health and capacity. The first one determines the safety of an EV, whereas the second one determines the cruise range. We provide labels for both tasks and elaborate on their definitions below. 
	
	Health determines whether a battery can operate safely without causing fire or sudden breakdowns. Hence our label on battery health focuses on battery system anomaly detection, aiming to identify any abnormal behavior or malfunctions in the battery system. In our case, an abnormal label is generated upon fire accidents and lithium plating reports. This health estimation task is crucial for ensuring the safety and reliability of EV batteries. By leveraging the dataset's rich collection of battery data, researchers and practitioners can develop and evaluate anomaly detection algorithms to enhance battery system diagnostics. The health labels are divided into two categories, with 1 indicating abnormalities. The distribution of labels is mentioned in Tabel~\ref{table:dataset_statistics}.
	
	\begin{figure}
		\centering
		\includegraphics[width=0.45\linewidth]{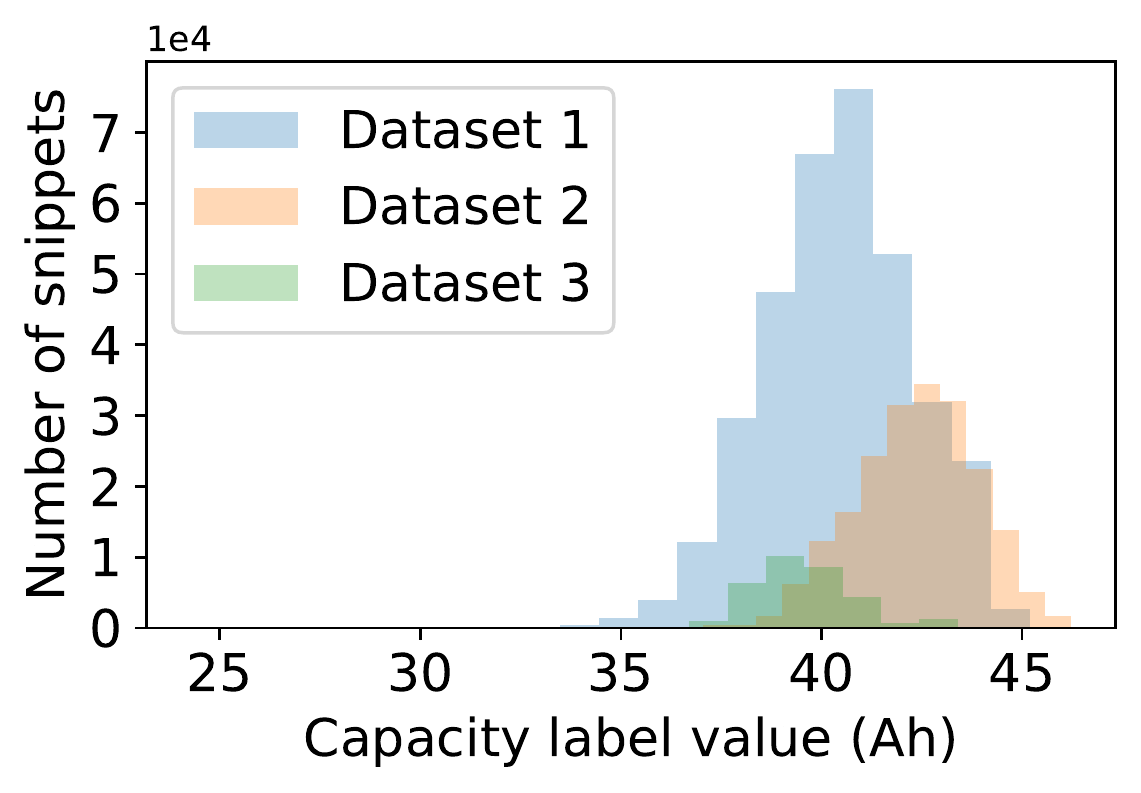}
		\caption{The distribution of the capacity label value of our dataset.}
		\label{fig:capacity_distribution}
	\end{figure}
	
	
	Capacity describes the amount of charge stored in a fully charged battery. This quantity correlates well with the cruise range of a vehicle. Hence, our second task is battery capacity estimation. Our label for capacity is defined as the amount of charge in $A \cdot h$ stored in the battery while the battery is charged from 3.77V to 4.05V with a 35A current level. Although this quantity can be computed directly with integration in a lab environment, estimating this can be highly nontrivial in practice because real-world vehicles do not follow a fixed charging protocol. This task contributes to improving the overall understanding and management of battery systems in electric vehicles. The capacity labels are real numbers, and Figure~\ref{fig:capacity_distribution} shows their distribution across three datasets. We can see that the distributions among the three manufacturers are slightly different.  
	

	Besides the above-mentioned two direct tasks, our dataset can also be used to develop transfer learning algorithms between different manufacturers and unsupervised or semi-supervised learning algorithms with partially labeled battery capacity labels. We welcome researchers to develop and validate other algorithms and applications using our dataset.
	Overall, the EVBattery dataset presents an opportunity to address these two important tasks in battery system anomaly detection and battery capacity estimation. Researchers can leverage this dataset to develop innovative solutions and techniques that enhance the performance, safety, and efficiency of EVs.
	
	
	Next, we show how predicting the battery health label can be viewed as an anomaly detection problem, and hence existing methods for anomaly / out-of-distribution detection can be applied.
	
	\subsection{Battery health as anomaly detection}
	
	Our first observation is that although our dataset is large with millions of charging snippets, the number of health labels is sparse for two reasons: first, battery failures are rare; second, the labels are on the vehicle level rather than on the charging level. For these reasons, direct supervised learning can easily lead to overfitting. 
	Consequently, a natural solution is to adopt the anomaly / out-of-distribution detection framework~\citep{AnomalydetectionAsurvey} and use unsupervised learning to avoid overfitting. We discuss some caveats in applying existing algorithms in the next paragraph.
	
	
	
	
	
	\paragraph{Difference to previous anomaly detection} Notice that the battery system failure labels are marked and can only be marked on the vehicle level rather than the snippet level. Adopting such a hierarchy labeling strategy is more practical for electric vehicles: it is impossible for expert battery engineers to judge whether the battery system is normal or not from the sequential data until a fire accident or other malfunction report. Such characteristic is one of the main differences from the previous event-based time series anomaly detection datasets. Event-based time series anomaly detection typically defines anomalies at specific time points or time intervals, such as aircraft system failures~\citep{wadidataset} or simulated attacks on water systems~\citep{wadidataset}. In the case of EV batteries, however, anomalies can only be defined at the level of individual vehicles. 
	Therefore, this task is different from previous time series anomaly detection tasks. An additional step to summarize predictions at the timestamp level into the vehicle level is required. 
	
\begin{figure}[h]
		\centering
		\subfloat[]{
			\label{fig:dataset_example1_a} 
			\includegraphics[width=0.35\linewidth]{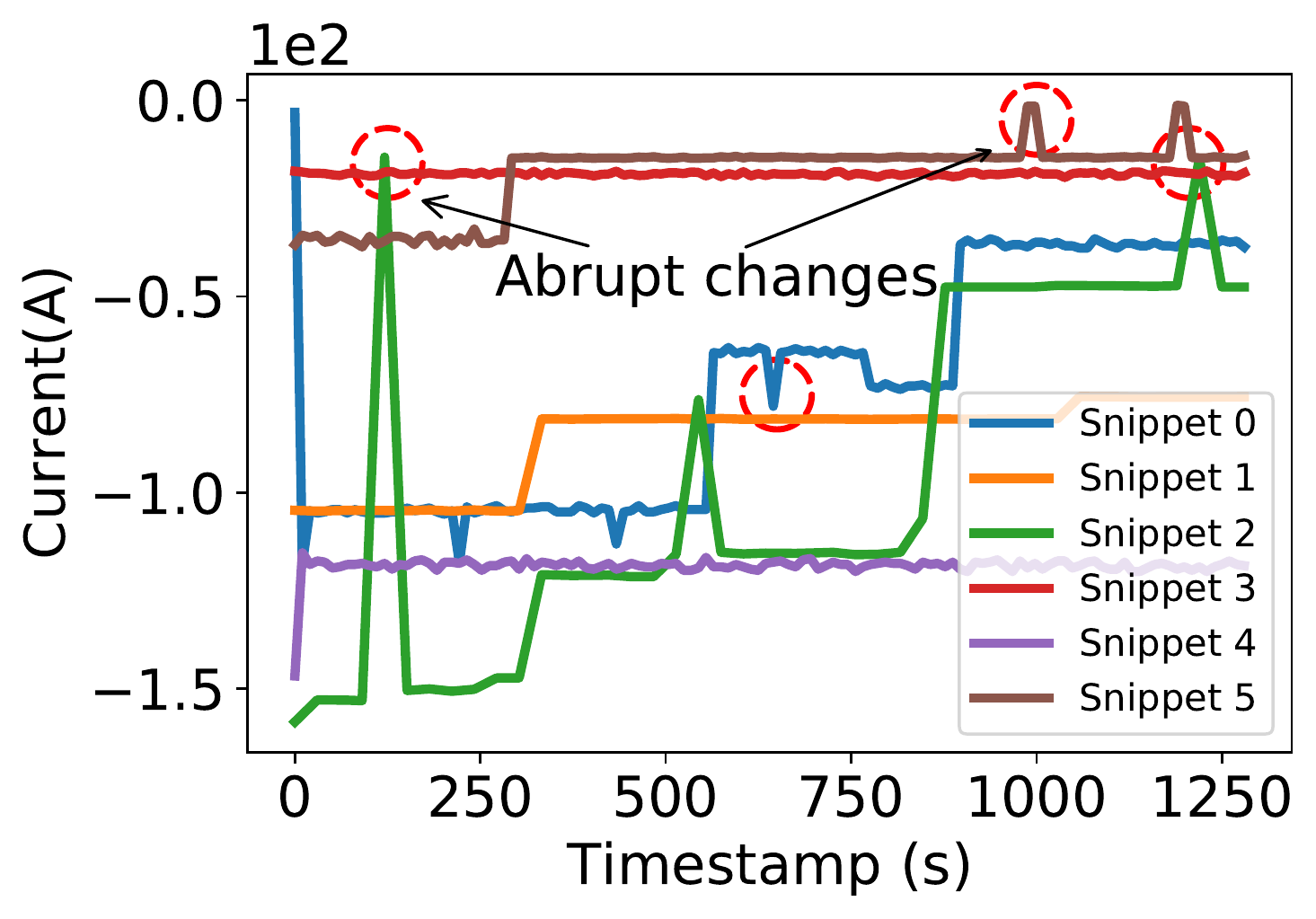}
		}
		\subfloat[]{
			\label{fig:dataset_example1_b} 
			\includegraphics[width=0.35\linewidth]{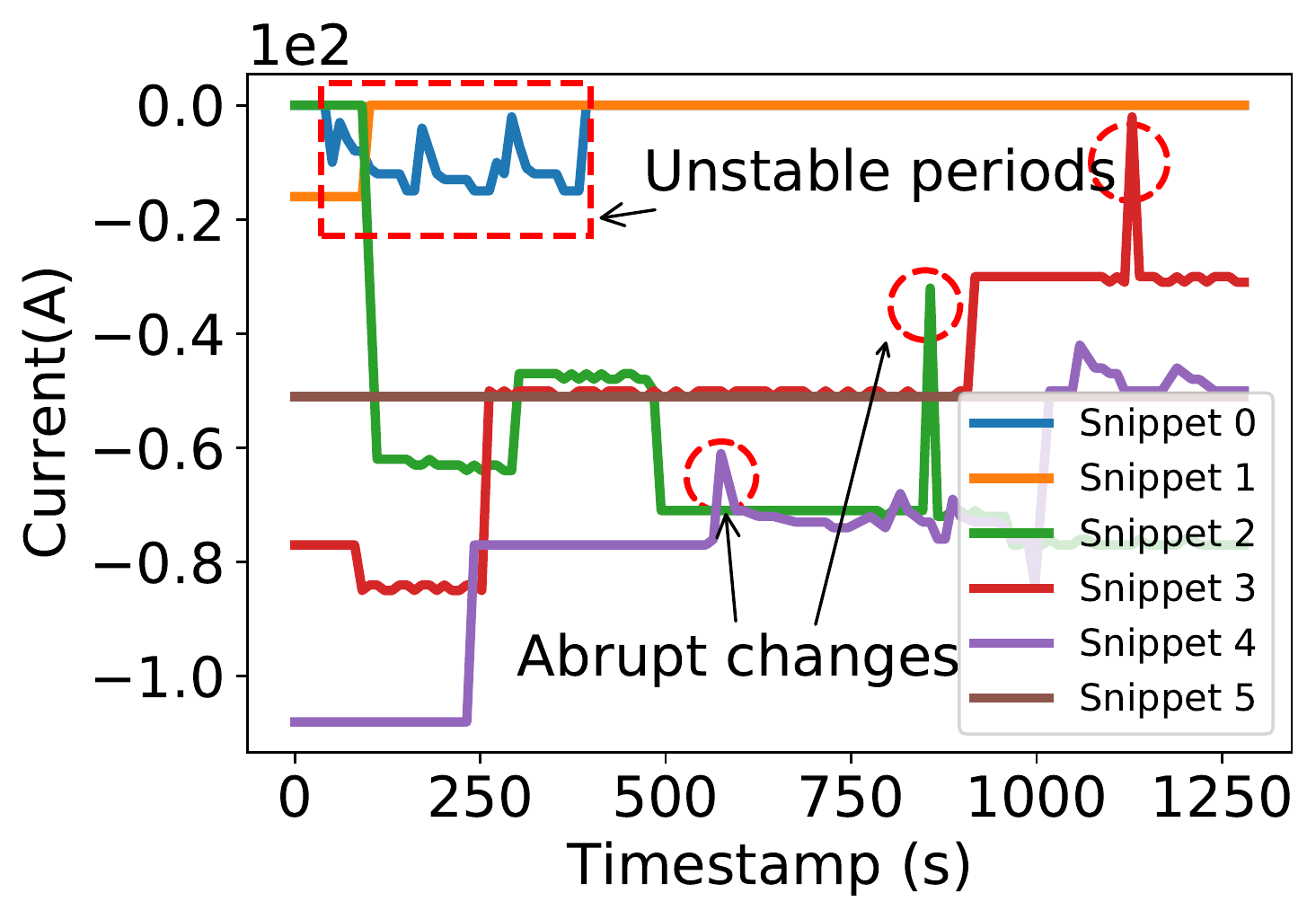}
		}
		\caption{Charging current patterns from different snippets at different charging phases. (a) Sampled normal vehicles' charging current patterns. (b) Sampled abnormal vehicles' charging current patterns. Abrupt changes and unstable charging periods are marked in both figures. These two figures illustrate the similarity of charging snippets from normal and fault systems, making anomaly detection difficult.}
		\label{fig:three_figures}
	\end{figure}

	\paragraph{Challenges in battery health anomaly detection} Here we discuss why traditional time-series algorithms would fail and highlight the importance of using deep learning methods to learn nonlinear high-level features for the health estimation task. Detecting anomalous batteries system is challenging. First, we observe that abrupt changes happen in both normal batteries and fault batteries. 
	Specifically, as Figure~\ref{fig:dataset_example1_a} shows, those snippets belong to normal vehicles not only have stable charging current records which meet the charging criteria but also have abrupt current changes during phase shifting~(green line) and constant current phase~(brown line). Meanwhile, those snippets belonging to fault vehicles in Figure~\ref{fig:dataset_example1_b} also have both cases, and some charging records~(blue line) have more violent fluctuations. Hence, utilizing the rarity of data patterns cannot easily distinguish normal vehicles from abnormal vehicles.
	
	\begin{figure}[htbp]
		\vspace{-0.5cm}
		\captionsetup[subfigure]{width=100pt}
		\centering
		\subfloat[]{
			\label{fig:dataset_distribution_0} 
			\includegraphics[width=0.30\linewidth]{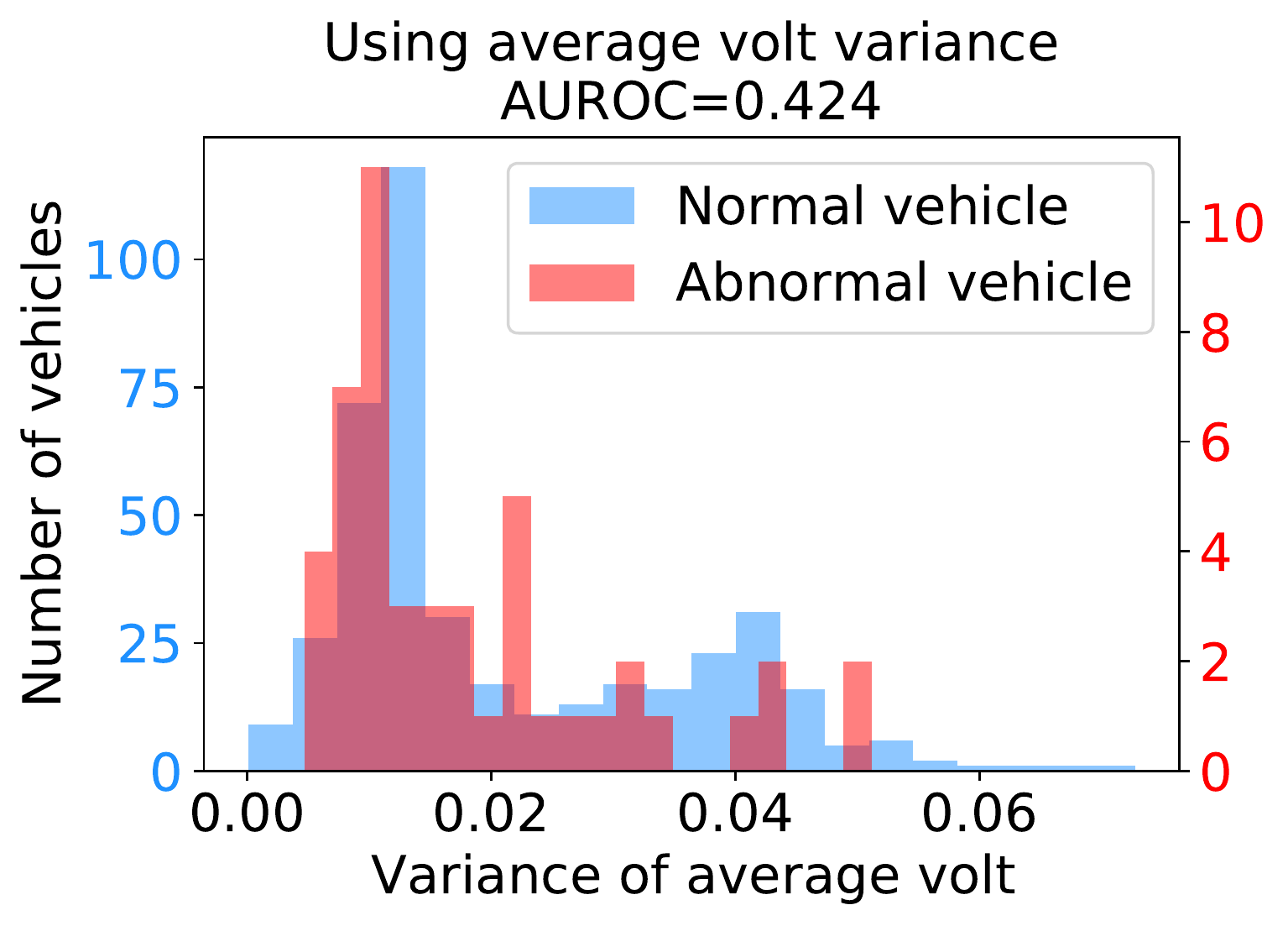}
		}
		\subfloat[]{
			\label{fig:dataset_distribution_1} 
			\includegraphics[width=0.30\linewidth]{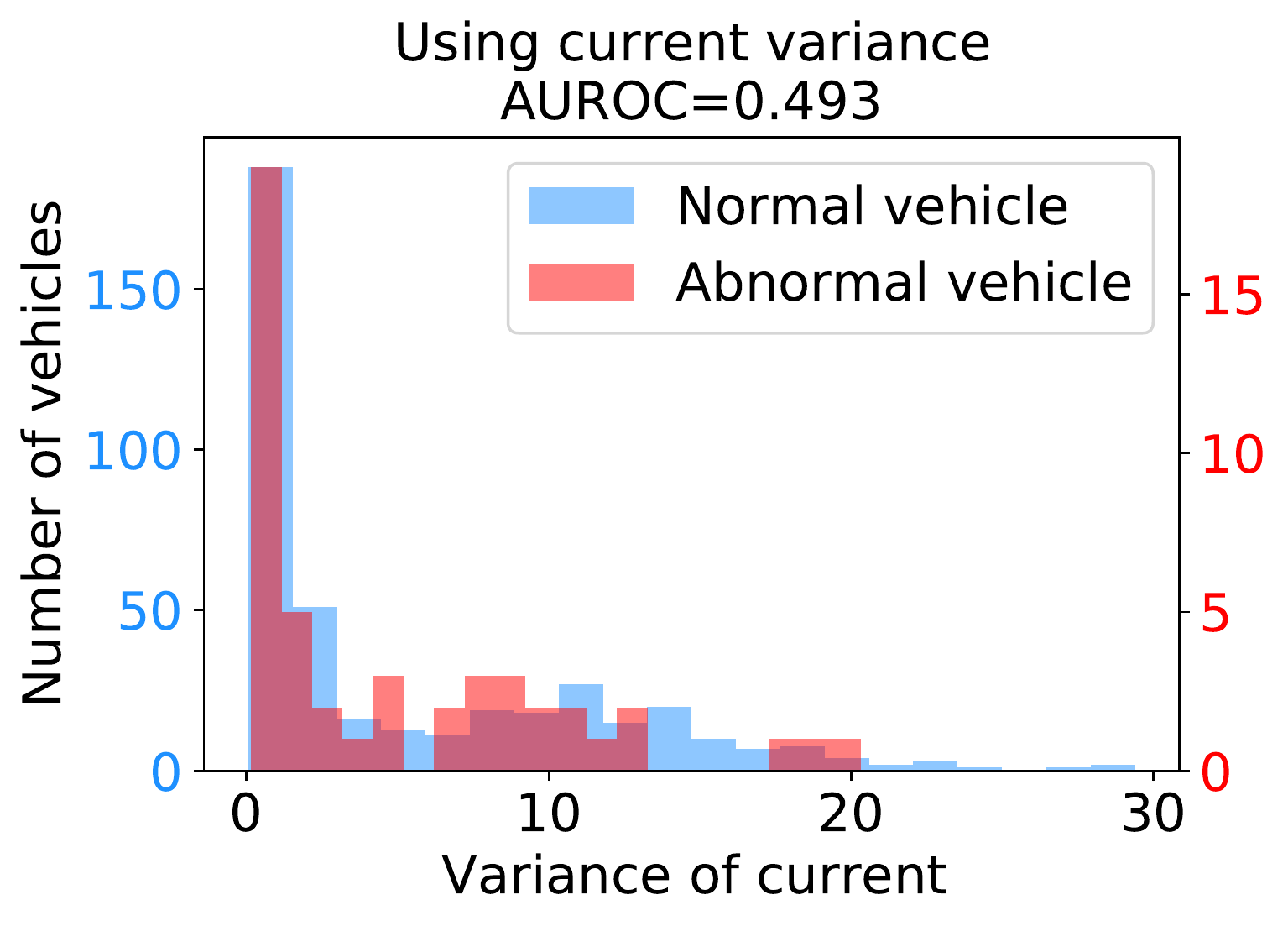}
		}
		\subfloat[]{
			\label{fig:dataset_distribution_2} 
			\includegraphics[width=0.30\linewidth]{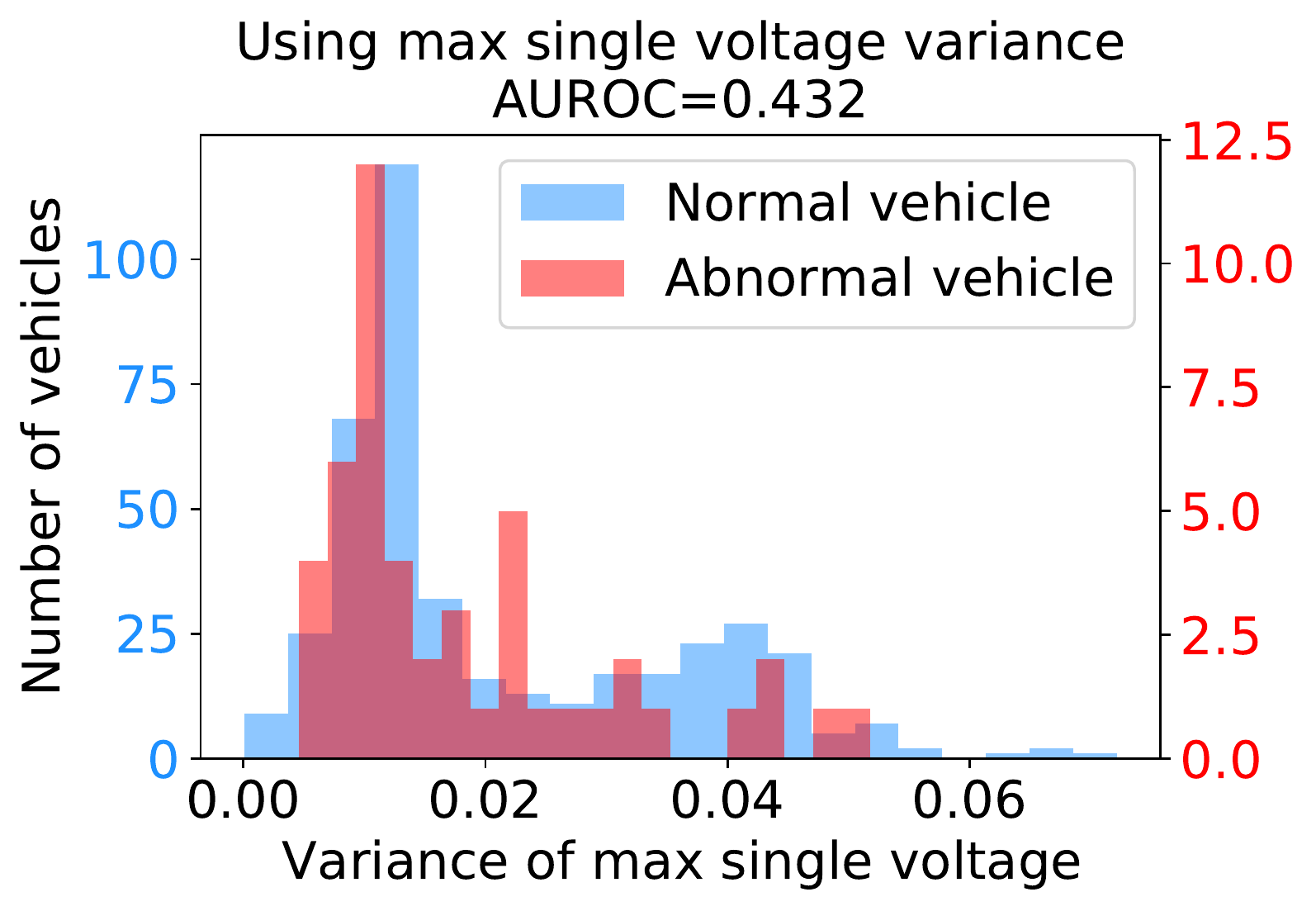}
		}\\
		\subfloat[]{
			\label{fig:dataset_distribution_3} 
			\includegraphics[width=0.30\linewidth]{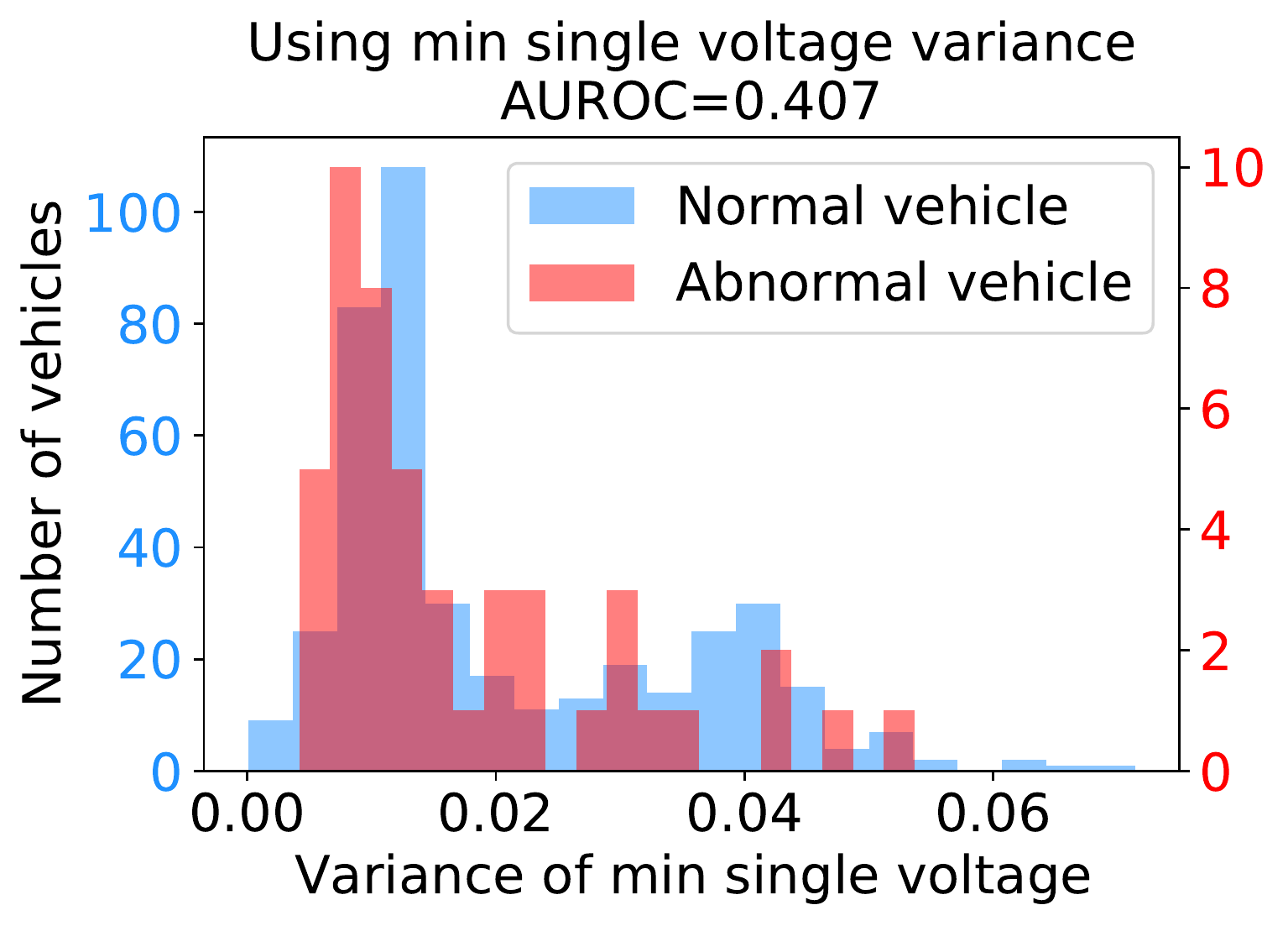}
		}
		\subfloat[]{
			\label{fig:dataset_distribution_4} 
			\includegraphics[width=0.30\linewidth]{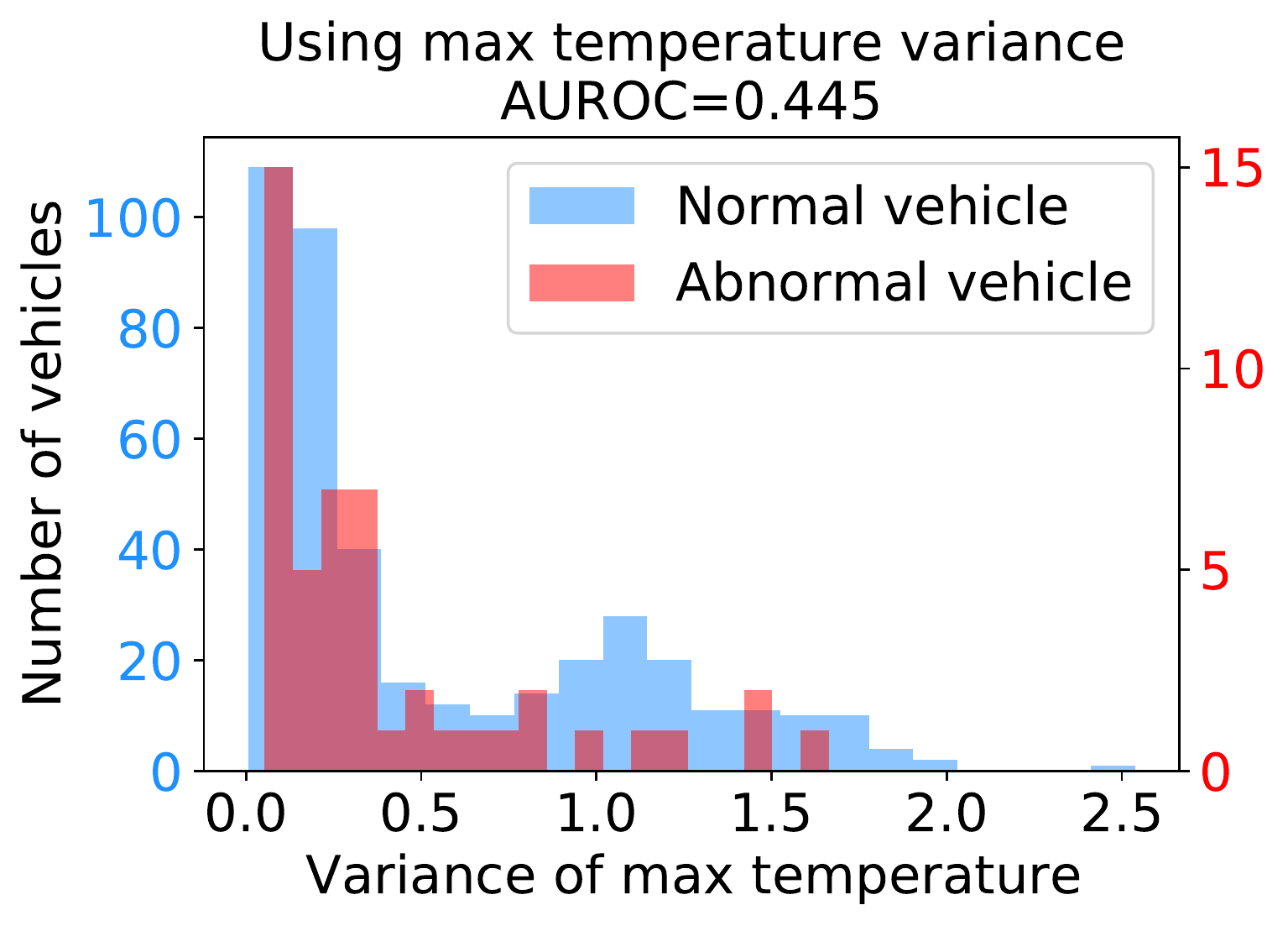}
		}
		\subfloat[]{
			\label{fig:dataset_distribution_5} 
			\includegraphics[width=0.30\linewidth]{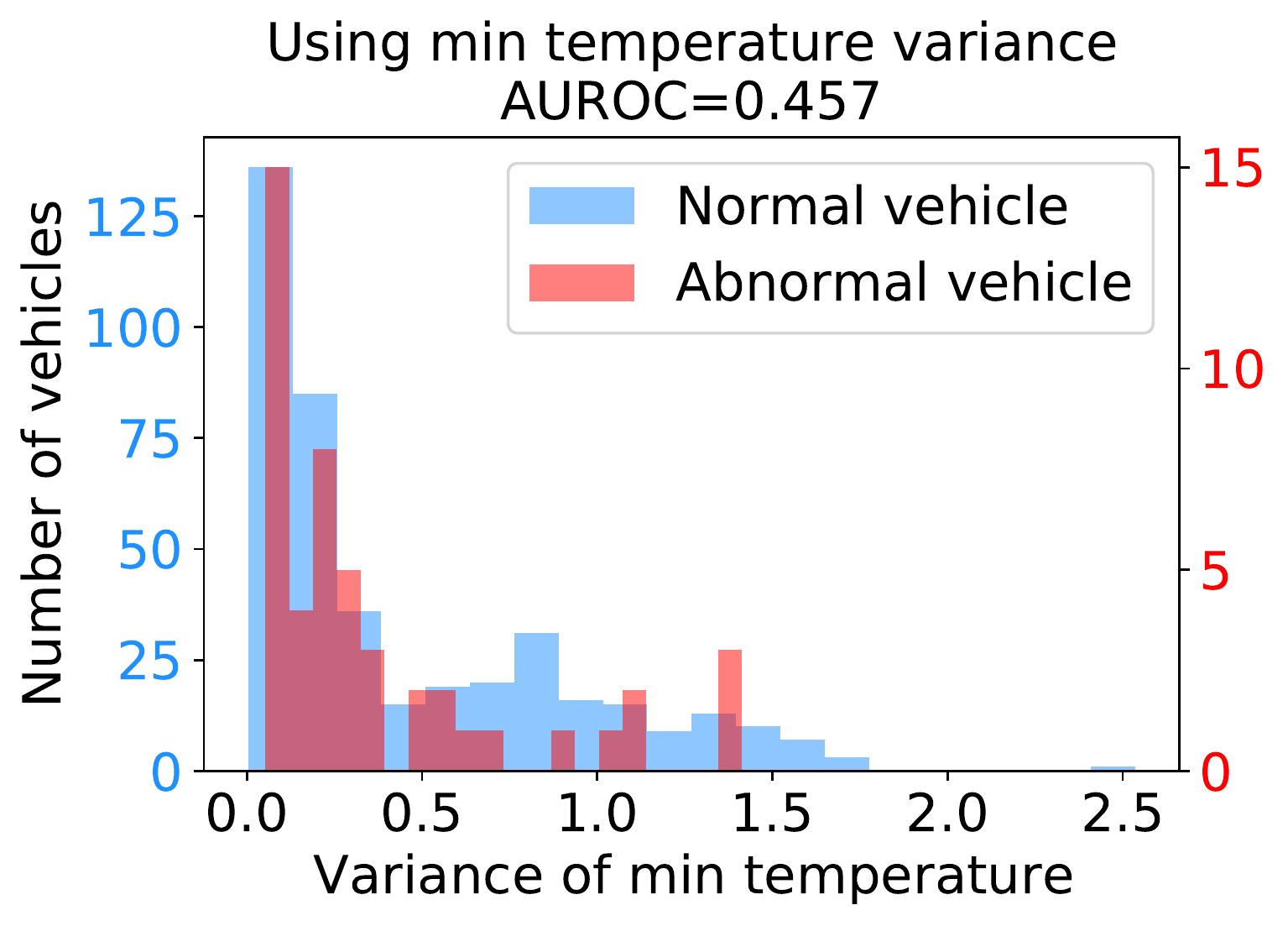}
		}
		\caption{AUROC values of using simple data statistics: (a) average variance of average volt (b) average variance of current (c) average variance of max single voltage (d) average variance of min single voltage (e) average variance of max temperature (f) average variance of min temperature. Results show that simple statistics can not distinguish battery system health status effectively.}
		\label{fig:variance_example1} 
	\end{figure}
	
	Second, we show that the variance of data has little predictive power. When it comes to vehicles with battery system failure problems, there is a tendency to assume that there is more variance in the time-series data for abnormal vehicles, or that abnormal vehicles have underrepresented charging records. Here we provide numerical results using the variance of data to differentiate normal battery systems from failure in Figure~\ref{fig:variance_example1}. The AUROC values indicate that such simple statistics are insufficient to detect battery system failure.

	We experimented with several commonly used time-series anomaly detection algorithms and a customized algorithm tailored to the characteristics of the dataset to observe their performance on the EVBattery dataset. The algorithms are adapted to our hierarchical time series anomaly detection task (see Appendix~\ref{appendix:adapt_to_timeood} for more details).
	These experiments allowed us to compare the performance of different algorithms and assess their suitability for anomaly detection in the EVBattery dataset. The results provided valuable insights for selecting the most effective algorithm, ultimately leading to the development of our DyAD algorithm.
	
	\subsection{Battery capacity estimation as regression}
	
	Battery capacity estimation is another crucial task in EV battery management. 
	By analyzing the battery data and charging records in the EVBattery dataset, researchers can explore the relationship between battery capacity and the charging process and develop machine learning and data analytic methods for capacity estimation. Such estimates can be used to monitor battery health, predict battery lifespan, and provide accurate references for driving and charging planning in electric vehicles.
	
	Research on the battery capacity estimation task using the EVBattery dataset can advance the field of electric vehicle battery management. Accurate battery capacity estimation contributes to improved range estimation, reduced battery degradation, and optimized charging and usage strategies, ultimately providing a better user experience. Battery capacity estimation is a standard regression problem in which the goal is to predict the remaining capacity of a battery based on various input features.
	
	
	\paragraph{Challenges in battery capacity estimation} Although Battery capacity estimation is a standard regression problem in which the goal is to predict the remaining capacity of a battery based on various input features, it has several challenges in practice. Due to the lack of uniformity in battery types and the complexity of operating conditions, battery capacity detection is challenging. The observed data is not obtained under standard laboratory conditions, further adding to the difficulty in precisely assessing battery capacity.
	
	Considering the large size of the dataset, some algorithms that are difficult to parallelize, such as linear regression and support vector regression, were not evaluated. Instead, we selected several tree-based algorithms and neural networks as benchmarks.
	Through this benchmarking process, we can gain insights into the strengths and weaknesses of different algorithms for battery capacity estimation. This information is valuable for selecting the most suitable algorithm for specific battery management applications and can guide further research and development in this field.
	
	We remark that those snippets without capacity labels are not used in our experiments. But they may be useful if researchers consider a semi-supervised case. We leave it as a future work.

	\section{Battery system anomaly detection}
	\label{sec:battery_system_anomaly_detection}
	In this section, we first introduce our proposed algorithm for this task and then benchmark several known algorithms. We show that estimating health can be challenging and research efforts specific to this problem can improve model performances.
	
	\subsection{A tailored algorithm for battery health anomaly detection} 
\begin{figure}[htbp]
		\centering
		\includegraphics
		[width=0.7\linewidth]
		{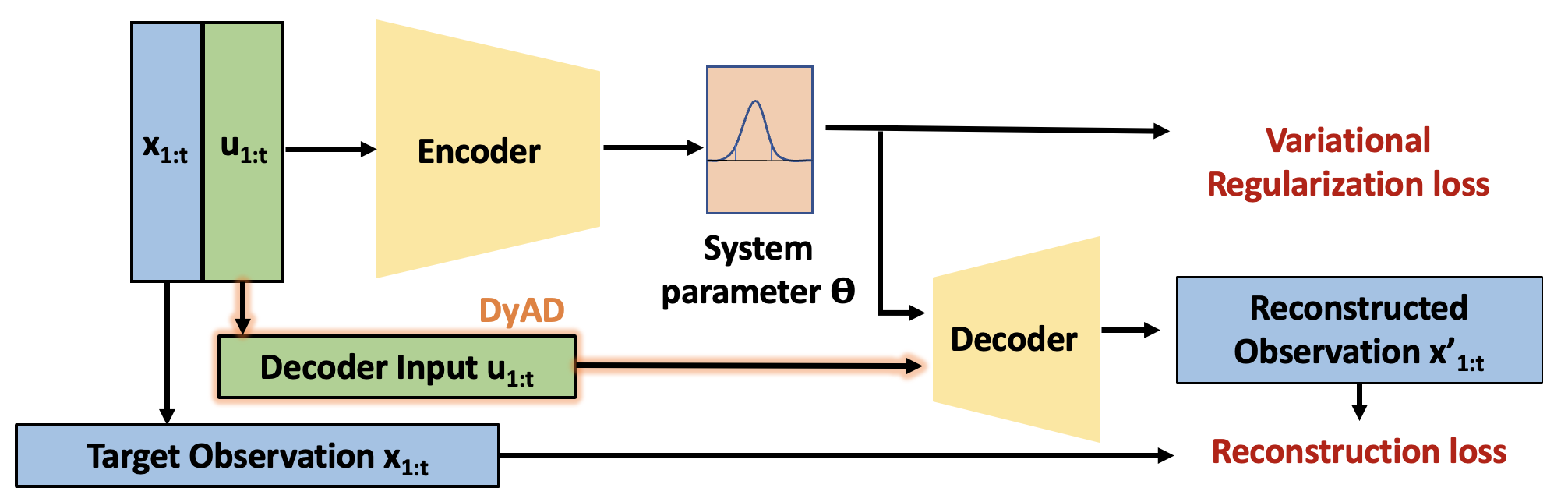}%
		\caption{The network architecture of DyAD. 
			The orange shade highlights the difference between DyAD and a standard VAE model.
		}\label{fig:train1}
	\end{figure}

Although we processed and organized the data such that the latest time series anomaly detection models~\cite{deepsvdd, DBLP:journals/corr/MalhotraRAVAS16, GDN} can be readily applied and tested, we highlight that the additional structure in EVBattery data leads to two major differences from most existing time series anomaly detection datasets. 
	First, rare signal patterns do not imply system anomaly. For example, if a charging current contains unexpected oscillation caused by faults at the charging station, the entire battery data should still be considered normal if the battery generates oscillating patterns as a healthy battery would generate. Taking advantage of this characteristic, we made improvements to the standard autoencoder algorithm and proposed the DyAD (Dynamic AutoDecoder) algorithm. The main difference is that the system input information is independently fed into the decoder, making it more suitable for capturing the system input and response dynamics during EV charging. See Figure~\ref{fig:train1} for an illustration of our model architecture. 
	Second, as the label belong to vehicles, snippet information may not truthfully reflect the health conditions of that snippet. To address this, we adopt a robust scoring procedure to build vehicle-level predictions from the associated charging snippets.
	The algorithm is shown in Alg.~\ref{alg:robust_score1}.
	In particular, we predict the abnormal degree of a charging snippet by thresholding the reconstruction error at value $\tau$ and then predict whether a vehicle is abnormal by averaging the top $h$ percentile errors. Both $\tau$ and $h$ are fine-tuned on the validation dataset. This procedure can be applied to all deep learning baselines.

	\begin{algorithm}[h]
		\caption{\label{alg:robust_score1}Pseudo code of the robust anomaly score} 
		\hspace*{0.02in}
		\hspace*{0.02in} {\bf Hyperparameters:} percentile $h$ and threshold $\tau$. \\
		\hspace*{0.02in} {\bf Input:} Charging snippet scores~$\Vec{r} = \{(r_i)\}_i$ from a vehicle. \\
		\hspace*{0.02in} {\bf Output:} 
		Vehicle-level prediction.
		\begin{algorithmic}[1]
			\State Sort $\Vec{r}$ by $r_i$ from large to small.
			\State Take the mean of the largest $h\%$ as the vehicle's score.
			\State Predict a vehicle as abnormal if the average of the largest $h\%$ is greater than $\tau$.
			
		\end{algorithmic}
		
	\end{algorithm}

	\subsection{Results}
	
	We evaluate our framework together with existing anomaly detection algorithms on the released large-scale dataset with RTX 2080Ti graphics cards. The considered algorithms include autoencoder~(AE)~\cite{autoencoderanomaly}, Deep SVDD~\cite{deepsvdd}, LSTMAD~\cite{DBLP:journals/corr/MalhotraRAVAS16}, MTAD-GAT~\cite{MTADGAT}, and GDN~\cite{GDN}.  The implementation details can be found in Appendix~\ref{appendix:model_details} and our released code. Following the common evaluation metric in anomaly detection with sparse labels, we use the area under the receiver operating characteristic~(AUROC) curves to show the results. 
	
	All experiments were conducted using five-fold cross-validation to compute average performance. Considering the scarcity of anomalous vehicles in real-world scenarios and to ensure algorithm fairness, we only utilized four folds of normal vehicle data during the training process of all models. When selecting the values of percentile parameter $h$ and threshold parameter $\tau$, we used one fold of anomaly vehicle data along with the training data. During testing, we used the remaining one fold of normal vehicle data and four folds of anomalous vehicle data.

	The interpolated averaged ROC curves and the mean and variance of AUROC values are shown in Figure~\ref{fig:avr_roc_robust} in Appendix~\ref{appendix_auroc_figure} and Table~\ref{table_roc1}, respectively. Battery dataset 2 is omitted since it has less than five abnormal vehicles. We can see that our proposed algorithm DyAD achieves the best performance on our large-scale dataset, demonstrating the effectiveness of adding prior for a specific dataset and the tailored model architecture in capturing anomalies in the EVBattery dataset. Additionally, although different algorithms showed improvement or degradation in performance when using datasets from a single manufacturer, our DyAD results demonstrate the potential to achieve better detection performance even when the data source distribution is more homogeneous. Meanwhile, this is the first time to deploy deep learning algorithms to detect electric vehicle battery system failure in such a large and real-world dataset. 
	
	\begin{table}[ht]
		\centering
		\caption{\label{table_roc1} Mean and standard variance of \emph{test} AUROC~(\%) values on all vehicles from three manufactuers. Among all the considered algorithms, DyAD achieves the best detection results. Bold denotes the best results. The ROC curve figures are shown in Appendix~\ref{appendix_auroc_figure}.}
		\resizebox{0.7\columnwidth}{!}{%
			\begin{tabular}{cccc}
				\toprule[1pt]
				Algorithm & Battery dataset 1 & Battery dataset 3 &All datasets \\ \midrule
				AE & $59.3\pm5.0$  & $64.5\pm7.0$  & $68.3\pm6.9$ \\ 
				Deep SVDD  &  $51.3\pm2.8$  &  $50.3\pm14.8$  &  $60.3\pm2.1$  \\ 
				LSTMAD & $55.7\pm4.8$ & $60.3\pm6.8$ &$59.9\pm4.0$ \\
				MTAD-GAT &  $50.6\pm4.7$   & $55.7\pm6.0$  &  $57.1\pm4.1$  \\ 
				GDN  &  $61.7\pm9.4$   & $61.8\pm6.3$  &  $57.8\pm3.5$  \\ 
				\textbf{DyAD}~(Ours) &  $\bm{77.9\pm5.0}$  &   $\bm{69.8\pm12.8}$ &   $\bm{70.0\pm3.5}$  \\ 
				\bottomrule
			\end{tabular}
		}
		\vspace{-0.4cm}
	\end{table}

	\vspace{-0.3cm}
	\paragraph{Remark} We noticed that an AUROC value of around 70\% for anomaly detection under-performs metrics in other datasets. In the field of image anomaly detection, the detection AUROC of many algorithms on the corresponding datasets has been able to reach 95\%~\cite{Ma, CSI}. On small-scale simple datasets for time series anomaly detection, some algorithms can also achieve F1 scores of 0.8$\sim$0.9~\cite{GDN, MTADGAT}. This indicates that the hierarchical time series anomaly detection task is much more challenging. It is possible to improve detection efficiency through engineering methods, such as incorporating abnormal vehicles into the training process. We leave different formulations of the health estimation task as future research topics.
	
	\vspace{-0.1cm}
	\section{Battery capacity estimation}
	\vspace{-0.1cm}
	\label{sec:battery_capacity_estimation}
	In this section, we present the results of the battery capacity estimation using various machine learning algorithms and neural networks. The goal of this experiment was to estimate the battery capacity based on the available data. We compared the performance of five different algorithms: Random Forest~\citep{ho1995random}, XGBoost~\citep{chen2016xgboost}, Multi-Layer Perceptron, Gated CNN~\citep{dauphin2017language}, and LSTM~\citep{hochreiter1997long}. The training details are in Appendix~\ref{appendix:train_detail_capacity}. Two traditional machine learning algorithms with parallel acceleration are presented in Appendix~\ref{appendix_parallel_ml_capacity_result}. 
	
	We use five-fold cross-validation similar to the battery system anomaly detection task. The difference is that we do not differentiate between normal and abnormal vehicles. Only snippets with capacity labels are included in the training and test set. Due to label sparsity, battery dataset 3 fails to construct the five-fold cross-validation so we omit it.
	To assess the accuracy of the battery capacity estimation, we employed the Root Mean Square Error (RMSE) metric. A lower RMSE value indicates a more accurate estimation of the battery capacity. Additionally, we also calculated the standard deviation of the RMSE to measure the consistency of the algorithm's performance.
	
	From the results in Table~\ref{table:capacity_estimation_rmse}, we observe that all five algorithms achieved relatively low RMSE values, indicating their ability to estimate the battery capacity accurately. Among the algorithms, LSTM exhibited the lowest RMSE value of 1.73 with a standard deviation of 0.06 on three datasets. It seems that LSTM was consistent in its capacity estimations, demonstrating its capability in capturing complex temporal patterns and providing accurate capacity estimations. As regression is a standard and straightforward task, we did not make any specific modifications to the algorithms based on the characteristics of the dataset. We welcome researchers to develop more effective algorithms on top of our baseline methods.

	\begin{table}[h]
		\centering
		\vspace{-0.4cm}
		\caption{\label{table:capacity_estimation_rmse}Mean and standard variance of \emph{test} RMSE values of battery 
			capacity estimation. 
		}
		\resizebox{0.72\columnwidth}{!}{%
			\begin{tabular}{cccc}
				\toprule
				& Battery dataset 1 & Battery dataset 2 & All datasets\\ \midrule
				Random Forest & $1.78\pm0.07$ & $1.42\pm0.11$ & $1.88\pm0.04$        \\
				XGBoost       & $1.74\pm0.06$ & \bm{$1.39\pm0.13$}   & $1.81\pm0.06$   \\
				MLP          & $1.74\pm0.08$ & $1.43\pm0.12$ & $1.78\pm0.03$        \\
				GCNN       & $1.77\pm0.06$ & $1.42\pm0.11$ & $1.74\pm0.03$   \\
				LSTM      & \bm{$1.69\pm0.06$} & $1.41\pm0.13$ & \bm{$1.73\pm0.06$}       \\
				\bottomrule
			\end{tabular}
			
		}
		\vspace{-0.6cm}
	\end{table}
	
	\section{Discussion}
	\vspace{-0.3cm}
	The tasks of our EVBattery dataset are not limited to EV battery health anomaly detection and capacity estimation. The dataset contains EV batteries from three manufacturers so the transfer learning task between different datasets can also be considered.
	In the task of battery health anomaly detection, we adopt an unsupervised learning method, that is, abnormal vehicles are not included in the training data. This is a challenging scenario. A slightly simpler scenario extension is to add abnormal vehicles to the training set, which becomes a classification task, or anomaly detection methods such as one class SVM can be used to achieve better detection results. In the battery capacity detection task, we choose simple supervised learning, that is, only using charging snippets labeled by capacity as training and testing data. It's an open problem of how to improve performance with snippets that don't have charging tags.
	
	\newpage

	\appendix

	\bibliographystyle{plainnat}
	\bibliography{example_paper}

\begin{thebibliography}{26}
\providecommand{\natexlab}[1]{#1}
\providecommand{\url}[1]{\texttt{#1}}
\expandafter\ifx\csname urlstyle\endcsname\relax
  \providecommand{\doi}[1]{doi: #1}\else
  \providecommand{\doi}{doi: \begingroup \urlstyle{rm}\Url}\fi

\bibitem[Aggarwal(2013)]{autoencoderanomaly}
Charu~C. Aggarwal.
\newblock \emph{Outlier Analysis}.
\newblock Springer, 2013.
\newblock ISBN 978-1-4614-6395-5.
\newblock \doi{10.1007/978-1-4614-6396-2}.
\newblock URL \url{https://doi.org/10.1007/978-1-4614-6396-2}.

\bibitem[Ahmed et~al.(2017)Ahmed, Palleti, and Mathur]{wadidataset}
Chuadhry~Mujeeb Ahmed, Venkata~Reddy Palleti, and Aditya~P. Mathur.
\newblock {WADI:} a water distribution testbed for research in the design of
  secure cyber physical systems.
\newblock In Panagiotis Tsakalides and Baltasar Beferull{-}Lozano, editors,
  \emph{Proceedings of the 3rd International Workshop on Cyber-Physical Systems
  for Smart Water Networks, CySWATER@CPSWeek 2017, Pittsburgh, Pennsylvania,
  USA, April 21, 2017}, pages 25--28. {ACM}, 2017.
\newblock \doi{10.1145/3055366.3055375}.
\newblock URL \url{https://doi.org/10.1145/3055366.3055375}.

\bibitem[Chandola et~al.(2009)Chandola, Banerjee, and
  Kumar]{AnomalydetectionAsurvey}
Varun Chandola, Arindam Banerjee, and Vipin Kumar.
\newblock Anomaly detection: {A} survey.
\newblock \emph{{ACM} Comput. Surv.}, 41\penalty0 (3):\penalty0 15:1--15:58,
  2009.
\newblock \doi{10.1145/1541880.1541882}.

\bibitem[Chen and Guestrin(2016)]{chen2016xgboost}
Tianqi Chen and Carlos Guestrin.
\newblock Xgboost: A scalable tree boosting system.
\newblock In \emph{Proceedings of the 22nd acm sigkdd international conference
  on knowledge discovery and data mining}, pages 785--794, 2016.

\bibitem[Dauphin et~al.(2017)Dauphin, Fan, Auli, and
  Grangier]{dauphin2017language}
Yann~N Dauphin, Angela Fan, Michael Auli, and David Grangier.
\newblock Language modeling with gated convolutional networks.
\newblock In \emph{International conference on machine learning}, pages
  933--941. PMLR, 2017.

\bibitem[Deng and Hooi(2021)]{GDN}
Ailin Deng and Bryan Hooi.
\newblock Graph neural network-based anomaly detection in multivariate time
  series.
\newblock In \emph{Thirty-Fifth {AAAI} Conference on Artificial Intelligence,
  {AAAI} 2021, Thirty-Third Conference on Innovative Applications of Artificial
  Intelligence, {IAAI} 2021, The Eleventh Symposium on Educational Advances in
  Artificial Intelligence, {EAAI} 2021, Virtual Event, February 2-9, 2021},
  pages 4027--4035. {AAAI} Press, 2021.
\newblock URL \url{https://ojs.aaai.org/index.php/AAAI/article/view/16523}.

\bibitem[Goebel et~al.(2008)Goebel, Saha, Saxena, Celaya, and
  Christophersen]{goebel2008prognostics}
Kai Goebel, Bhaskar Saha, Abhinav Saxena, Jose~R Celaya, and Jon~P
  Christophersen.
\newblock Prognostics in battery health management.
\newblock \emph{IEEE instrumentation \& measurement magazine}, 11\penalty0
  (4):\penalty0 33--40, 2008.

\bibitem[Ho(1995)]{ho1995random}
Tin~Kam Ho.
\newblock Random decision forests.
\newblock In \emph{Proceedings of 3rd international conference on document
  analysis and recognition}, volume~1, pages 278--282. IEEE, 1995.

\bibitem[Hochreiter and Schmidhuber(1997)]{hochreiter1997long}
Sepp Hochreiter and J{\"u}rgen Schmidhuber.
\newblock Long short-term memory.
\newblock \emph{Neural computation}, 9\penalty0 (8):\penalty0 1735--1780, 1997.

\bibitem[Hong et~al.(2019)Hong, Wang, and Yao]{hong2019fault}
Jichao Hong, Zhenpo Wang, and Yongtao Yao.
\newblock Fault prognosis of battery system based on accurate voltage abnormity
  prognosis using long short-term memory neural networks.
\newblock \emph{Applied Energy}, 251:\penalty0 113381, 2019.

\bibitem[Lee et~al.(2018)Lee, Lee, Lee, and Shin]{Ma}
Kimin Lee, Kibok Lee, Honglak Lee, and Jinwoo Shin.
\newblock A simple unified framework for detecting out-of-distribution samples
  and adversarial attacks.
\newblock In \emph{NeurIPS 2018}, 2018.

\bibitem[Li et~al.(2020)Li, Zhang, Liu, Wang, and Zhang]{li2020battery}
Da~Li, Zhaosheng Zhang, Peng Liu, Zhenpo Wang, and Lei Zhang.
\newblock Battery fault diagnosis for electric vehicles based on voltage
  abnormality by combining the long short-term memory neural network and the
  equivalent circuit model.
\newblock \emph{IEEE Transactions on Power Electronics}, 36\penalty0
  (2):\penalty0 1303--1315, 2020.

\bibitem[Liu and Xu(2020)]{liu2020data}
Wei Liu and Yan Xu.
\newblock Data-driven online health estimation of li-ion batteries using a
  novel energy-based health indicator.
\newblock \emph{IEEE Transactions on Energy Conversion}, 35\penalty0
  (3):\penalty0 1715--1718, 2020.

\bibitem[Malhotra et~al.(2016)Malhotra, Ramakrishnan, Anand, Vig, Agarwal, and
  Shroff]{DBLP:journals/corr/MalhotraRAVAS16}
Pankaj Malhotra, Anusha Ramakrishnan, Gaurangi Anand, Lovekesh Vig, Puneet
  Agarwal, and Gautam Shroff.
\newblock Lstm-based encoder-decoder for multi-sensor anomaly detection.
\newblock \emph{CoRR}, abs/1607.00148, 2016.
\newblock URL \url{http://arxiv.org/abs/1607.00148}.

\bibitem[Mauler et~al.(2021)Mauler, Duffner, Zeier, and
  Leker]{mauler2021battery}
Lukas Mauler, Fabian Duffner, Wolfgang~G Zeier, and Jens Leker.
\newblock Battery cost forecasting: a review of methods and results with an
  outlook to 2050.
\newblock \emph{Energy \& Environmental Science}, 2021.

\bibitem[Ng et~al.(2020)Ng, Zhao, Yan, Conduit, and Seh]{ng2020predicting}
Man-Fai Ng, Jin Zhao, Qingyu Yan, Gareth~J Conduit, and Zhi~Wei Seh.
\newblock Predicting the state of charge and health of batteries using
  data-driven machine learning.
\newblock \emph{Nature Machine Intelligence}, 2\penalty0 (3):\penalty0
  161--170, 2020.

\bibitem[Radford et~al.(2019)Radford, Wu, Child, Luan, Amodei, Sutskever,
  et~al.]{radford2019language}
Alec Radford, Jeffrey Wu, Rewon Child, David Luan, Dario Amodei, Ilya
  Sutskever, et~al.
\newblock Language models are unsupervised multitask learners.
\newblock \emph{OpenAI blog}, 1\penalty0 (8):\penalty0 9, 2019.

\bibitem[Rezvanizaniani et~al.(2014)Rezvanizaniani, Liu, Chen, and
  Lee]{rezvanizaniani2014review}
Seyed~Mohammad Rezvanizaniani, Zongchang Liu, Yan Chen, and Jay Lee.
\newblock Review and recent advances in battery health monitoring and
  prognostics technologies for electric vehicle (ev) safety and mobility.
\newblock \emph{Journal of power sources}, 256:\penalty0 110--124, 2014.

\bibitem[Ruff et~al.(2018)Ruff, G{\"{o}}rnitz, Deecke, Siddiqui, Vandermeulen,
  Binder, M{\"{u}}ller, and Kloft]{deepsvdd}
Lukas Ruff, Nico G{\"{o}}rnitz, Lucas Deecke, Shoaib~Ahmed Siddiqui, Robert~A.
  Vandermeulen, Alexander Binder, Emmanuel M{\"{u}}ller, and Marius Kloft.
\newblock Deep one-class classification.
\newblock In Jennifer~G. Dy and Andreas Krause, editors, \emph{Proceedings of
  the 35th International Conference on Machine Learning, {ICML} 2018,
  Stockholmsm{\"{a}}ssan, Stockholm, Sweden, July 10-15, 2018}, volume~80 of
  \emph{Proceedings of Machine Learning Research}, pages 4390--4399. {PMLR},
  2018.
\newblock URL \url{http://proceedings.mlr.press/v80/ruff18a.html}.

\bibitem[Schmuch et~al.(2018)Schmuch, Wagner, H{\"o}rpel, Placke, and
  Winter]{schmuch2018performance}
Richard Schmuch, Ralf Wagner, Gerhard H{\"o}rpel, Tobias Placke, and Martin
  Winter.
\newblock Performance and cost of materials for lithium-based rechargeable
  automotive batteries.
\newblock \emph{Nature Energy}, 3\penalty0 (4):\penalty0 267--278, 2018.

\bibitem[Tack et~al.(2020)Tack, Mo, Jeong, and Shin]{CSI}
Jihoon Tack, Sangwoo Mo, Jongheon Jeong, and Jinwoo Shin.
\newblock {CSI:} novelty detection via contrastive learning on distributionally
  shifted instances.
\newblock In Hugo Larochelle, Marc'Aurelio Ranzato, Raia Hadsell,
  Maria{-}Florina Balcan, and Hsuan{-}Tien Lin, editors, \emph{Advances in
  Neural Information Processing Systems 33: Annual Conference on Neural
  Information Processing Systems 2020, NeurIPS 2020, December 6-12, 2020,
  virtual}, 2020.
\newblock URL
  \url{https://proceedings.neurips.cc/paper/2020/hash/8965f76632d7672e7d3cf29c87ecaa0c-Abstract.html}.

\bibitem[Widodo et~al.(2011)Widodo, Shim, Caesarendra, and
  Yang]{widodo2011intelligent}
Achmad Widodo, Min-Chan Shim, Wahyu Caesarendra, and Bo-Suk Yang.
\newblock Intelligent prognostics for battery health monitoring based on sample
  entropy.
\newblock \emph{Expert Systems with Applications}, 38\penalty0 (9):\penalty0
  11763--11769, 2011.

\bibitem[Xue et~al.(2021)Xue, Li, Zhang, Shen, Chen, and Liu]{xue2021fault}
Qiao Xue, Guang Li, Yuanjian Zhang, Shiquan Shen, Zheng Chen, and Yonggang Liu.
\newblock Fault diagnosis and abnormality detection of lithium-ion battery
  packs based on statistical distribution.
\newblock \emph{Journal of Power Sources}, 482:\penalty0 228964, 2021.

\bibitem[Yang et~al.(2020)Yang, Xiong, Ma, and Lin]{yang2020characterization}
Ruixin Yang, Rui Xiong, Suxiao Ma, and Xinfan Lin.
\newblock Characterization of external short circuit faults in electric vehicle
  li-ion battery packs and prediction using artificial neural networks.
\newblock \emph{Applied Energy}, 260:\penalty0 114253, 2020.

\bibitem[Zhao et~al.(2020)Zhao, Wang, Duan, Huang, Cao, Tong, Xu, Bai, Tong,
  and Zhang]{MTADGAT}
Hang Zhao, Yujing Wang, Juanyong Duan, Congrui Huang, Defu Cao, Yunhai Tong,
  Bixiong Xu, Jing Bai, Jie Tong, and Qi~Zhang.
\newblock Multivariate time-series anomaly detection via graph attention
  network.
\newblock In Claudia Plant, Haixun Wang, Alfredo Cuzzocrea, Carlo Zaniolo, and
  Xindong Wu, editors, \emph{20th {IEEE} International Conference on Data
  Mining, {ICDM} 2020, Sorrento, Italy, November 17-20, 2020}, pages 841--850.
  {IEEE}, 2020.
\newblock \doi{10.1109/ICDM50108.2020.00093}.
\newblock URL \url{https://doi.org/10.1109/ICDM50108.2020.00093}.

\bibitem[Zheng et~al.(2020)Zheng, Lu, Gao, Han, Feng, and
  Ouyang]{zheng2020micro}
Yuejiu Zheng, Yifan Lu, Wenkai Gao, Xuebing Han, Xuning Feng, and Minggao
  Ouyang.
\newblock Micro-short-circuit cell fault identification method for lithium-ion
  battery packs based on mutual information.
\newblock \emph{IEEE Transactions on Industrial Electronics}, 68\penalty0
  (5):\penalty0 4373--4381, 2020.

\end{thebibliography}
	
	\section*{Checklist}

	\begin{enumerate}
		
		\item For all authors...
		\begin{enumerate}
			\item Do the main claims made in the abstract and introduction accurately reflect the paper's contributions and scope?
			\answerYes{}
			\item Did you describe the limitations of your work?
			\answerYes{}
			\item Did you discuss any potential negative societal impacts of your work?
			\answerYes{}
			\item Have you read the ethics review guidelines and ensured that your paper conforms to them?
			\answerYes{}
		\end{enumerate}
		
		\item If you are including theoretical results...
		\begin{enumerate}
			\item Did you state the full set of assumptions of all theoretical results?
			\answerNA{}
			\item Did you include complete proofs of all theoretical results?
			\answerNA{}
		\end{enumerate}
		
		\item If you ran experiments (e.g. for benchmarks)...
		\begin{enumerate}
			\item Did you include the code, data, and instructions needed to reproduce the main experimental results (either in the supplemental material or as a URL)?
			\answerYes{See Section~\ref{sec:evbattery_dataset} and Appendix~\ref{appendix:more_dataset_information}, \ref{appendix:model_details}, and~\ref{appendix:train_detail_capacity}.}
			\item Did you specify all the training details (e.g., data splits, hyperparameters, how they were chosen)?
			\answerYes{See Section~\ref{sec:battery_system_anomaly_detection}, \ref{sec:battery_capacity_estimation} and Appendix~\ref{appendix:model_details}, and~\ref{appendix:train_detail_capacity}.}
			\item Did you report error bars (e.g., with respect to the random seed after running experiments multiple times)?
			\answerYes{Variance of the results are reported.}
			\item Did you include the total amount of compute and the type of resources used (e.g., type of GPUs, internal cluster, or cloud provider)?
			\answerYes{See Section~\ref{sec:battery_system_anomaly_detection}.}
		\end{enumerate}
		
		\item If you are using existing assets (e.g., code, data, models) or curating/releasing new assets...
		\begin{enumerate}
			\item If your work uses existing assets, did you cite the creators?
			\answerYes{}
			\item Did you mention the license of the assets?
			\answerYes{}
			\item Did you include any new assets either in the supplemental material or as a URL?
			\answerYes{}
			\item Did you discuss whether and how consent was obtained from people whose data you're using/curating?
			\answerYes{}
			\item Did you discuss whether the data you are using/curating contains personally identifiable information or offensive content?
			\answerYes{}
		\end{enumerate}
		
		\item If you used crowdsourcing or conducted research with human subjects...
		\begin{enumerate}
			\item Did you include the full text of instructions given to participants and screenshots, if applicable?
			\answerNA{}
			\item Did you describe any potential participant risks, with links to Institutional Review Board (IRB) approvals, if applicable?
			\answerNA{}
			\item Did you include the estimated hourly wage paid to participants and the total amount spent on participant compensation?
			\answerNA{}
		\end{enumerate}
		
	\end{enumerate}
	
	
	\appendix
	
	\section{More dataset information}
	\label{appendix:more_dataset_information}
	
\begin{figure}
		\centering
		\includegraphics[width=8cm]{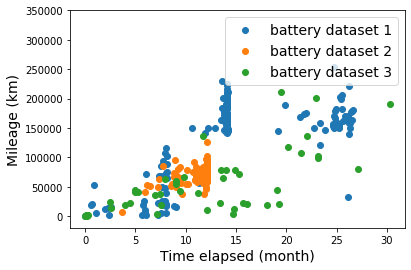}
		\caption{\label{figure_collection_time}The mileage information vs. elapsed collection time for three datasets.}
		\label{fig:enter-label}
	\end{figure}

	
	In this section, we offer an additional explanation of the dataset and code. The provided download link \url{https://1drv.ms/f/s!AnE8BfHe3IOlg13v2ltV0eP1-AgP?e=9o4zgL} includes three compressed battery datasets and a compressed package containing code. The usage of the dataset and code is under a CC BY-NC-SA license. The data are saved as pickle files which can be loaded with Pytorch. The code package includes instructions on how to load the data, generate a PyTorch standard dataset, select either the complete dataset or data from a specific manufacturer, and all the algorithm implementations mentioned in our experiments, including our proposed DyAD algorithm. The mileage and collection duration are shown in Figure~\ref{figure_collection_time}.
	
	The dataset and code link will be maintained through OneDrive.  We promise the long-term accessibility of the dataset. The DOI of our dataset is 10.6084/m9.figshare.23301881. 
	
	\section{Training details of the battery health anomaly detection task}
	\label{appendix:model_details}
	The implementation details can be found in our released code. We partially use official and public code resources from PyOD\footnote{\url{https://github.com/yzhao062/pyod}}, LSTMAD\footnote{\url{https://github.com/PyLink88/Recurrent-Autoencoder}} and GDN\footnote{\url{https://github.com/d-ailin/GDN}}. 
	
	\subsection{Autoencoder}
	
	The network adopts an encoder-decoder structure with batch normalization layers, drop out layers, and sigmoid activation functions. The latent dimensions are [64, 32, 32, 64] in the encoder and decoder. We train the network 10 epochs with a batch size of 128. We use the Adam optimizer with a learning rate of 0.001.

	\subsection{Deep SVDD}
	The feature extraction network of the deep SVDD is an autoencoder with hidden dimensions [64, 32, 32, 64] with sigmoid activation functions. The SVDD loss is computed on the middle 32-dimensional latent feature. We also adopt the reconstruction loss to help the network learn better data representation. We train 10 epochs with a batch size of 64 using the Adam optimizer with a learning rate of 0.001. 
	
	\subsection{LSTMAD}
	The LSTMAD uses the LSTM layers and LSTM cells to encode and decode the input data, respectively. The latent feature dimension is 32. We use an Adam optimizer with a learning rate of 0.001 and train 20 epochs with a batch size of 128. The reconstruction loss function is the mean absolute error.  
	
	\subsection{MTAD-GAT}
	For the spacecraft dataset, the detection results are directly from the official paper~\cite{MTADGAT}. For the battery dataset, we set the window length to 100 and train the graph network 30 epochs with a batch size of 256. Only the feature dimension number is modified. The other parameters are set to the default value.

	\subsection{GDN}
	Following \cite{GDN}, we adapt the graph layers to fit our input data. Stacked fully-connected layers are attached at the end of the graph layer. The latent dimension is 128. The window length of the time-series data is set to 32 for the battery dataset and 128 for the spacecraft datasets. We use an Adam optimizer with a learning rate of 0.001 to train 20 epochs with a batch size of 128. 
	
	\subsection{DyAD}
	For the battery dataset, we use GRU as the recurrent unit and train the network 10 epochs with a cosine annealing Adam optimizer. The hidden size and latent size are set to 64 and 32, respectively. 
	
	\subsection{Running time and resources}
	
	In Table~\ref{table_running_time}, we show the running time of the battery system anomaly detection task. All experiments are run on a machine with 4 RTX 2080Ti graphic cards. 
	
	\begin{table}[!ht]
		\centering
		\caption{\label{table_running_time}Algorithm running time. }
		\begin{tabular}{|l|l|l|l|l|l|l|}
			\hline
			~ & AE & Deep SVDD & LSTMAD & MTAD-GAT & GDN & DyAD \\ \hline
			Training time (s) & 354 & 221 & 3320 & 2242 & 1780 & 3786 \\ \hline
			Testing time (s) & 25 & 16 & 245 & 92 & 308 & 322 \\ \hline
		\end{tabular}
	\end{table}
	
	\section{Results on unanonymized data}
	
	\label{appendix_unanomymized_result}
	
	The datasets we release publicly are anonymized at the request of our data providers. In addition, we conducted the same experiment internally on unanonymized data, and the results are shown in Table~\ref{table_battery_health_unanonymized} and Table~\ref{table_battery_capacity_unanonymized}. It can be seen that the results of each algorithm are not significantly different from their performance on anonymized data, which shows that the anonymization process does not contaminate the information of battery health and capacity estimation.
	
	\begin{table}[!ht]
		\centering
		\caption{\label{table_battery_health_unanonymized}Battery health anomaly detection results on unanonymized data.}
		\begin{tabular}{|l|l|l|l|l|}
			\hline
			~ & Battery dataset 1 & Battery dataset 3 & All datasets & \begin{tabular}{c}
				Average difference to\\ anonymized results  
			\end{tabular} \\ \hline
			AE & 59.8$\pm$4.7 & 64.7$\pm$6.3 & 68.2$\pm$6.6 & +0.20 \\ \hline
			Deep SVDD & 52.0$\pm$2.8 & 50.8$\pm$12.0 & 60.6$\pm$1.4 & +0.50 \\ \hline
			LSTMAD & 55.5$\pm$3.9 & 60.1$\pm$6.8 & 60.2$\pm$4.1 & -0.03 \\ \hline
			MTAD-GAT & 50.4$\pm$5.0 & 55.0$\pm$5.4 & 56.8$\pm$4.1 & -0.40 \\ \hline
			GDN & 61.8$\pm$9.7 & 61.8$\pm$5.2 & 57.6$\pm$3.4 & -0.03 \\ \hline
			DyAD & \bm{$78.0\pm5.1$} & \bm{$69.6\pm11.2$} & \bm{$70.0\pm3.7$} & -0.03 \\ \hline
		\end{tabular}
	\end{table}
	
	\vspace{-15px}
	
	\begin{table}[!ht]
		\centering
		\caption{\label{table_battery_capacity_unanonymized}Battery capacity estimation results on unanomymized data.}
		\begin{tabular}{|l|l|l|l|l|}
			\hline
			~ & Battery dataset 1 & Battery dataset 3 & All datasets & \begin{tabular}{c}
				Average difference to\\ anonymized results  
			\end{tabular} \\ \hline
			\begin{tabular}{c}
				Random\\ Forest
				
			\end{tabular} & 1.77$\pm$0.07 & 1.42$\pm$0.10 & 1.89$\pm$0.05 & 0 \\ \hline
			XGBoost & 1.75$\pm$0.06 & \bm{$1.39\pm0.12$} & 1.81$\pm$0.06 & 0.003 \\ \hline
			MLP & 1.74$\pm$0.07 & 1.42$\pm$0.12 & 1.79$\pm$0.03 & 0 \\ \hline
			GCNN & 1.77$\pm$0.07 & 1.42$\pm$0.11 & \bm{$1.73\pm0.04$} & -0.003 \\ \hline
			LSTM & \bm{$1.69\pm0.06$} & 1.40$\pm$0.13 & \bm{$1.73\pm0.05$} & -0.003 \\ \hline
		\end{tabular}
	\end{table}
	
	\section{Adapt to hierarchical time series anomaly detection}
	\label{appendix:adapt_to_timeood}
	Different from previous time series anomaly detection algorithms where anomalies are defined at specific time points or time intervals, our hierarchical time-series task needs to aggregate information from all charging snippets belonging to a vehicle and then make a decision. More specifically, the reconstruction error or other anomaly scores are gathered from algorithm training. To obtain the vehicle anomaly scores, an immediate idea to obtain the anomaly score of a vehicle is to take the average of its associated charging snippets. However, this hasty approach overlooks a fact: even a faulty car does not necessarily mean that all its charging segments in the past look abnormal. On the contrary, many anomalies are transient, which means that charging segments of an anomalous car may only contain a small portion that appears abnormal. This actually inspired our proposed robust scoring function~\ref{alg:robust_score1}, which selects a subset of segments as a decision-making basis by adjusting the threshold.
	
	\section{Training details of the battery capacity estimation task}
	\label{appendix:train_detail_capacity}
	The LSTM neural network consists of an LSTM layer followed by two fully connected layers with ReLU activation functions. The hidden dimension is set to 32. The network is trained using the Adam optimizer with a learning rate of 0.001 for 10 epochs. Mean square error is used as the training objective. The two tree algorithms, XGBoost and random forest, use root mean square error as the training loss. The depth is set to be 4. The hidden size of the multi-layer perceptron is 32 and we use the same training optimization strategy as LSTM. The Gated CNN in our experiments consists of four layers of the gated structure followed by a fully connected layer. The residual link is activated among gated layers. The same training optimization strategy is used. 
	
	\section{Parallel machine learning algorithms on capacity estimation}
	
	\label{appendix_parallel_ml_capacity_result}
	
	We also implemented two machine learning algorithms for the capacity estimation task. In order to improve training efficiency, we parallelized these two algorithms. The results are shown in Table~\ref{table_parallel_machine_learning}. Two traditional algorithms are implemented with no performance enhancement. No obvious effect improvement was observed.
	
	\begin{table}[!ht]
		\centering
		\caption{\label{table_parallel_machine_learning}Additional experiments for battery capacity estimation. }
		\begin{tabular}{|l|l|l|l|}
			\hline
			~ & Battery dataset 1 & Battery dataset 2 & All datasets \\ \hline
			Parallel SVR & 1.78$\pm$0.06 & 1.46$\pm$0.17 & 1.87$\pm$0.07 \\ \hline
			Parallel linear regression & 1.78$\pm$0.07 & 1.48 $\pm$0.16 & 1.91$\pm$0.07 \\ \hline
		\end{tabular}
	\end{table}
	
	\section{More experiment figures}
	
	\label{appendix_auroc_figure}
	
\begin{figure}[htbp]
		\centering
		\includegraphics[width=0.45\linewidth]{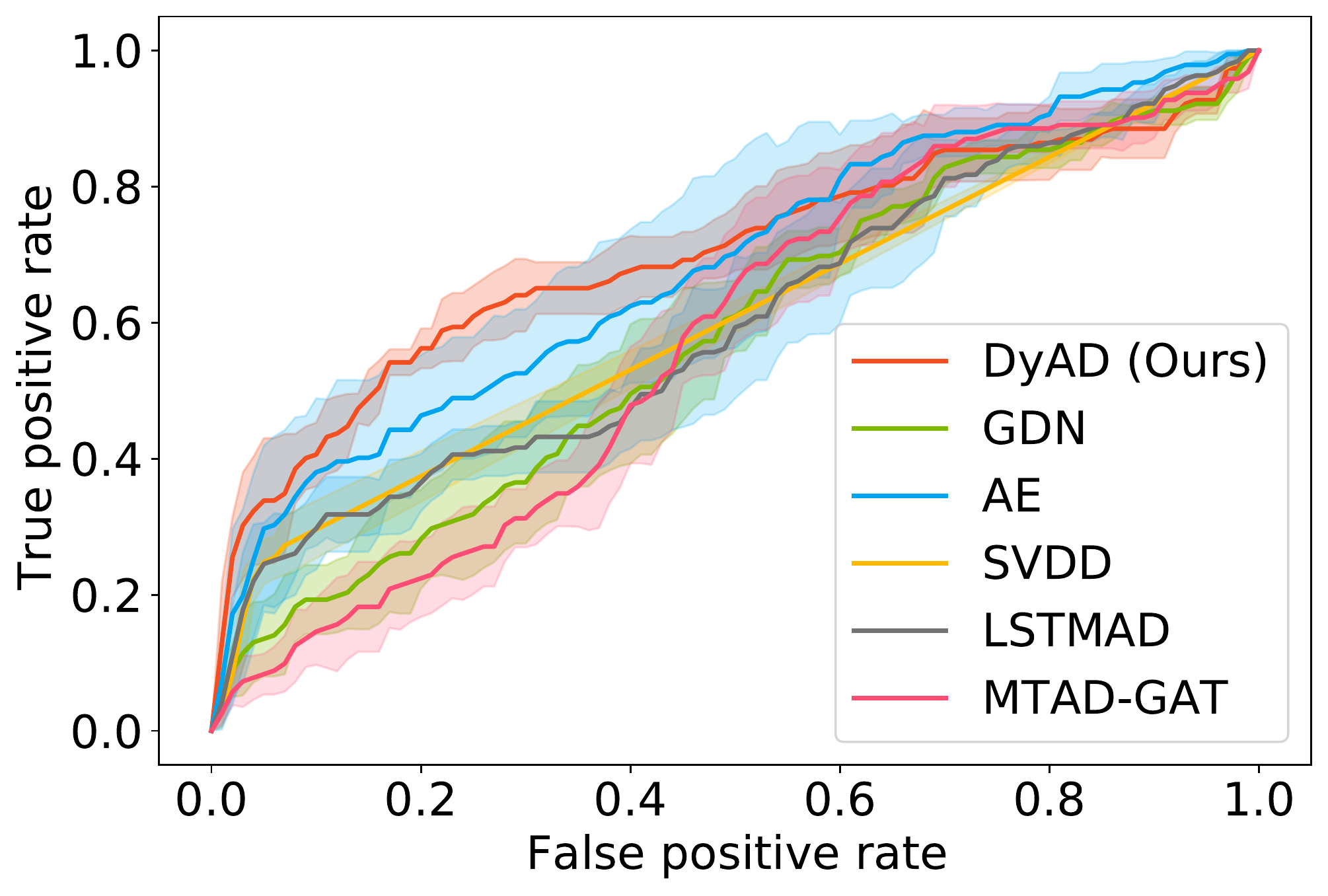}
		\caption{\label{fig:avr_roc_robust}Interpolated averaged ROC curves of several algorithms on three datasets. The shaded area represents the five-fold variance.}
	\end{figure}

	Unlike AUROC, which reflects the overall performance of a model, the ROC curve provides a more detailed comparison of varying true positive rate and false positive rate. The ROC curves are shown in Figure~\ref{fig:avr_roc_robust}.

	
\end{document}